%% file: iclr2025_conference.tex
\documentclass{article} 
\usepackage{tccml_iclr2025_conference,times}

\input{math_commands.tex}

\usepackage{hyperref}
\usepackage[capitalise]{cleveref}
\usepackage{url}
\usepackage{siunitx}
\usepackage{graphicx}
\usepackage{booktabs} 
\usepackage{multirow}

\title{Exploring Design Choices for Autoregressive Deep Learning Climate Models}

\author{Florian Gallusser, Simon Hentschel, Anna Krause, Andreas Hotho\\
Data Science Chair, 
Center for Artificial Intelligence and Data Science (CAIDAS), \\
University of Würzburg \\
\texttt{\{gallusser,hentschel,anna.krause, hotho\}@informatik.uni-wuerzburg.de} \\
}

\iclrfinalcopy 
\begin{document}

\maketitle

\begin{abstract}

Deep Learning (DL) models have achieved state-of-the-art performance in medium-range weather prediction (MWP) but often fail to maintain physically consistent rollouts beyond 14 days. In contrast, a few atmospheric models demonstrate stability over decades, though the key design choices enabling this remain unclear. This study quantitatively compares the long-term stability of three prominent DL-MWP architectures — FourCastNet, SFNO, and ClimaX — trained on ERA5 reanalysis data at \SI{5.625}{\degree} resolution. We systematically assess the impact of autoregressive training steps, model capacity, and choice of prognostic variables, identifying configurations that enable stable 10-year rollouts while preserving the statistical properties of the reference dataset. Notably, rollouts with SFNO exhibit the greatest robustness to hyperparameter choices, yet all models can experience instability depending on the random seed and the set of prognostic variables. \footnote{Code is available at \url{https://github.com/LSX-UniWue/dl-climate-models}}

\end{abstract}

\section{Introduction}

Over the past few years autoregressive Deep Learning (\textit{DL}) models have emerged that are en par with physics-based state-of-the-art medium range weather prediction systems while only requiring a fraction of the computational costs for inference \citep{lam2023a, bi2023a, price2025}. Given an initial state of the atmosphere, those models iteratively predict future states for a set of prognostic variables, which is commonly referred to as autoregressive rollout. However, the forecasts of popular weather models tend to degrade beyond 14 days and result in prediction of system states with physically unplausible distributions \citep{karlbauer2024, mccabe2023, chattopadhyay2023}. This instability poses a fundamental challenge when considering DL models for earth system simulations, where interactions between atmosphere, oceans, land, and ice are simulated over climate timescales, i.e., decades to hundreds of years. While simulations with physics-based earth system models are a foundational scientific tool to understand and project past, present, and future climate change \cite{eyring2016b}, reducing their immense computational costs while maintaining physical accuracy is a key motivation for exploring DL-based alternatives.

Several DL approaches have demonstrated the ability to generate stable autoregressive trajectories for atmospheric dynamics over at least one year, leveraging either ERA5 reanalysis data \citep{hersbach2020} or outputs from atmospheric general circulation models (\textit{GCMs}). Most of these methods employ specialized grid representations or architectures to mitigate distortions introduced by the regular latitude-longitude grid \citep{weyn2020, karlbauer2024, bonev2023}. Also, the complexity of the emulated GCM \citep{scher2019} or the selection of variables from the reanalysis dataset seem to play an important role for stability, since proposed models are often trained on a very limited set of prognostic variables \citep{guan2024, weyn2020, cresswell-clay2024}. However, different forward stepping schemes have been employed, such as predicting only the next timestep \citep{bonev2023, karlbauer2024a} or predicting two timesteps at the same time \citep{weyn2020, karlbauer2024, cresswell-clay2024}, which were both combined with multiple autoregressive training steps.
The inclusion of top-of-the-atmosphere incident solar radiation as an external forcing variable has also been claimed to be a crucial factor for long-term stability \citep{guan2024}.
While first works that provided boundary conditions from the ocean just used annually repeating sea surface temperature \citep{watt-meyer2023, cachay2024}, recent efforts have also successfully coupled the atmospheric models to basic data-driven ocean models \citep{wang2024, cresswell-clay2024}. Despite these advancements, the fundamental design principles underpinning long-term stability remain relatively unclear, as existing studies primarily focus on demonstrating and evaluating stable models rather than systematically analyzing the factors that ensure stability.

A preliminary comparison of the long-term behavior of common weather model architectures was conducted by \citet{karlbauer2024a}, who demonstrated that even models lacking explicit spherical geometry and trained to predict a single future state per forward pass can achieve stable rollouts when trained on ERA5 data at a coarse resolution of \SI{5.625}{\degree}. While their results qualitatively indicate stability over 50-year rollouts, a rigorous quantitative assessment is still missing.

Building on this foundation, our study quantitatively evaluates the long-term stability of three prominent DL architectures: FourCastNet, SFNO, and ClimaX. We systematically investigate fundamental design choices, including the number of autoregressive steps during training, model capacity, and the choice of prognostic variables. Using ERA5 reanalysis data at \SI{5.625}{\degree} resolution, we identify model configurations that allow stable 10-year rollouts while preserving the statistical properties of the reference dataset. Moreover, we highlight the geometry-aware SFNO architecture as the least sensitive to hyperparameter variations, yet we observe that all models can experience severe instability for some training random seeds.

\section{Experimental Setup}

\paragraph{Dataset}
For our study, we use the Weatherbench1 dataset \citep{rasp2020} at a horizontal resolution of \SI{5.625}{\degree} (64x32 grid points) and at a temporal resolution of 6 hours. Weatherbench1 was created by bilinear interpolation to a regular latitude-longitude grid from the ERA5 reanalysis dataset \citep{hersbach2020} and is available at an hourly temporal resolution covering the years 1979 to 2018. We train our models on data from 1979 to 2007, use the year 2008 for validation during training and keep 10 years from 2009 to 2018 for testing. During our experiments, we use two distinct sets of prognostic variables (see Appendix \cref{tab:variables}). The first set contains 33 prognostic variables based on \cite{nguyen2023a} minus relative humidity and the level \SI{50}{\hecto\pascal}; the second set contains the eight prognostic variables from \cite{karlbauer2024a}. As a forcing variable, we use the top-of-the-atmosphere incident solar radiation (\textit{TISR}) computed from the total solar irradiance from the recommended solar forcing datataset \citep{matthes2017} for CMIP6\footnote{The code for computing TISR was adapted from the GraphCast repository \url{https://github.com/google-deepmind/graphcast}}. We employ z-score normalization to normalize each variable at each level by its global mean and standard deviation computed over the entire training set.

\paragraph{Models}
We experiment with three model architectures $f_\theta$: (i) \textbf{ClimaX} \citep{nguyen2023a} is based on a vision-transformer \citep{dosovitskiy2020} and employs a special variable tokenization and aggregation scheme, which was designed to use multiple physical quantities as input data. (ii) \textbf{FourCastNet} (\textit{FCN}) \citep{pathak2022} is based on the Adaptive Fourier Neural Operator \citep{guibas2021} which is an efficient token mixer in frequency space that allows FCNs to scale efficiently to large spatial resolutions. (iii) \textbf{Spherical Neural Operators} (\textit{SFNO}) \citep{bonev2023} generalizes Fourier Neural Operators \citep{li2020b} to the sphere by replacing the discrete Fourier transform by the spherical harmonics transform.

\paragraph{Training}
We compute the loss for a single training example with initial condition at timestep $t$, given a tensor of $K_p$ prognostic variables $X_t\in\mathbb{R}^{K_p\times W\times H}$, a tensor of $K_c$ of constant variables $C\in \mathbb{R}^{K_c\times W\times H}$ and $K_f$ initial forcing variables $F_t\in\mathbb{R}^{K_f\times W\times H}$ as
\begin{equation*}
    \mathcal{L}_t = \sum_{m=0}^{M-1} \mathcal{L}_t^m = \sum_{m=0}^{M-1}\sum_{k=1}^{K_p}\sum_{h=1}^H a_h\sum_{w=1}^{W} (X_{t+m\cdot\Delta t} + f_\theta(X_{t+m\cdot\Delta t},F_{t+m\cdot\Delta t},C) - X_{t+(m+1)\cdot\Delta t})^2
\end{equation*}
where $M$ is the number of autoregressive steps, $a_h$ is the area weight accounting for the area distortion introduced in the regular latitude-longitude grid of size $(W,H)$ and $\Delta t$ is equal to 6 hours. We sample initial conditions at \texttt{06:00}, \texttt{12:00}, \texttt{18:00} and \texttt{00:00} UTC for all days in the training period which results in \num{42368} training samples. Note that gradients are only computed after summing the loss $\mathcal{L}_t^m$ for all $M$-steps.

\paragraph{Experiment}
We perform a grid search over the following parameters: $f_\theta \in$ \{ClimaX, FCN, SFNO\}, the number of prognostic variables $K_p\in\{8,33\}$, the number of steps $M= \{1,2,4\}$ during training, the number of layers $L = \{4,6,8\}$ and the size of the hidden dimension $D = \{128, 256, 512\}$. We train each configuration with 10 different random seeds. Other hyperparameters were chosen based on \citet{karlbauer2024a} and are documented in the Appendix in \cref{tab:hyperparameters}.

\paragraph{Evaluation}
After training, we initialize each model on the 1st of January 2009 \texttt{00:00} UTC, the first timestep of the test dataset, and compute an autoregressive rollout for 10 years.
We compare the average over time of the predicted prognostic variables with the average over time of the target for the 10-year time period as a measure for long-term stability. As comparison metrics, we use the area-weighted mean squared error (RMSE). To obtain a single score for each model, we compare normalized predictions and targets, and average the per variable RMSE across the commonly evaluated variables \texttt{tas}, \texttt{uas}, \texttt{vas}, \texttt{ta850} and \texttt{zg500}. We refrain to rollout all configurations from multiple initial conditions since we observed in preliminary experiments that this does not effect the long-term behavior of the rollout. We consider climatology, i.e. the temporal average over the 29 years of the training data set, as the lower bound baseline method. The RMSE of the climatology indicates how much climate changed from the training period compared to the evaluation period, and since our model is not forced by essential climate drivers such as greenhouse gas emissions and the state of the ocean, we can not expect to outperform climatology with our models.

\section{Results \& Discussion}

\begin{figure}[h]
\centering
\includegraphics[width=\textwidth]{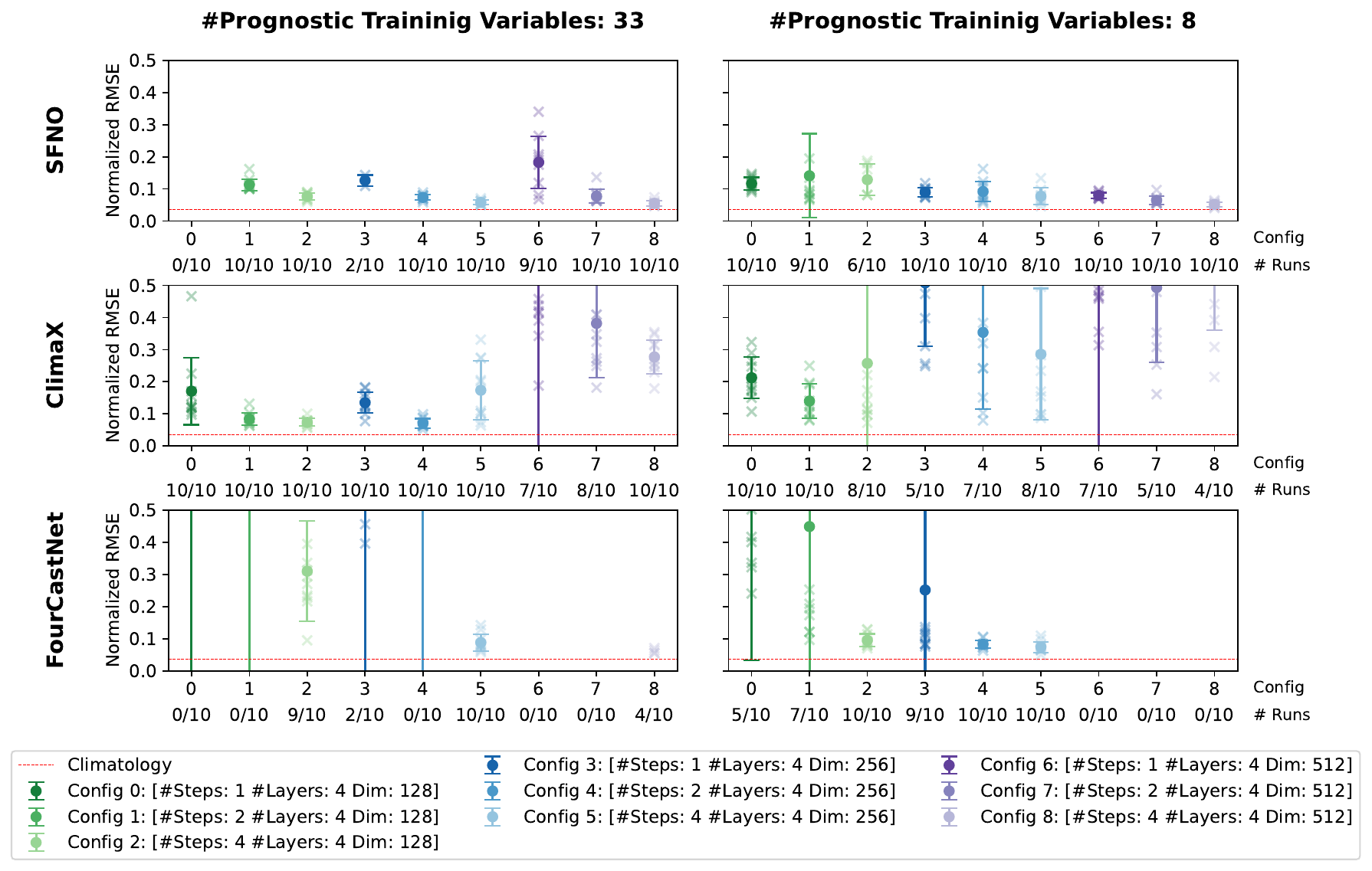}
\caption{Area-weighted normalized RMSE scores averaged over the variables \texttt{tas}, \texttt{uas}, \texttt{vas}, \texttt{ta850} and \texttt{zg500} for model configurations $0-8$ with layer size $L=4$. The mean (dots) and standard deviation (error bars) were computed across 10 seeds (crosses) with finite RMSE. For better readability, the y-axis is cut-off at 0.5 and the number of displayed runs out of 10 is shown on the x-axis.}
\label{fig:overview-layers4}
\end{figure}

We present the area-weighted normalized RMSE scores averaged across the evaluated variables for all trained models with a fixed number of layers $L=4$ in \cref{fig:overview-layers4}. However, the results discussed here generalize to configurations with $L=6$ and $L=8$ layers (see Appendix \cref{fig:overview-layers6} and \cref{fig:overview-layers8}).

Our results reveal that all model architectures can yield 10-year rollouts with comparably low RMSE scores close to the RMSE of climatology (see \cref{fig:overview-layers4}). However, across all configurations, the total number of SFNO models with RMSE score close to climatology is substantially higher than for FCN and ClimaX.
We verified that those rollouts do not only match well the temporal mean of the reference period, but also the temporal standard deviation (see Appendix \cref{fig:std-layers6}). Hence, those models preserve important distributional characteristics of the climate they were trained for.

Generally, we observe that multiple autoregressive steps during training reduces the RMSE of the 10-year rollouts. 
This effect is very dominant for the SFNO trained on 33 prognostic variables: multi-step training is much more stable than one-step training.
Also, FCN and ClimaX configurations with RMSE scores near climatology are all trained with a multi-step objective. However, multi-step training alone is not sufficient to obtain stable rollouts and choosing the right capacity of the model is crucial as well. ClimaX and FCN only achieve RMSE scores close to climatology, when trained with the small $D=128$ or medium $D=256$ hidden size across all number of layers. Contrarily, for SFNO models with $L=4$ layers, we see a clear trend that increasing the hidden dimension decreases the RMSE score. We hypothesize that implicitly satisfied geometrical constraints in the SFNO architecture allow to increase the model's capacity while FCN and ClimaX overfit to noisy signals close to the poles when capacity is increased.

Furthermore, we see that even for configurations that achieve RMSE scores close to climatology, some random seeds lead to severe instability of the 10-year rollout, e.g., SFNO trained on 8 prognostic variables, with the 4-step training objective and with hidden dimension 256. We verified that this can not be traced back to a decrease in 1-step prediction accuracy.
Additionally, we highlight that model configurations that are stable for one set of prognostic variables become unstable for the other set of prognostic variables. Interestingly, rollouts with ClimaX tend to become more stable with the larger set of prognostic variables, while contrarily the RMSE of rollouts with SFNO and FCN explodes with the larger set of prognostic variables for some configurations. 

\paragraph{Discussion}
While our results validate and quantitatively extend the findings of \citet{karlbauer2024a} that also non-geometry-aware architectures can achieve long stable rollouts with minimal drift, we showed that stability of rollouts with SFNO are much less sensitive to the choice of hyperparameters. Our results further confirm that multi-step training, which has been shown to mitigate autoregressive error growth in other experimental settings such as weather prediction \citep{lam2023a}, hybrid numerical-DL climate models \citep{kochkov2024}, and PDE learning \citep{brandstetter2021}, also enhances long-term stability in climate projections. Also \citet{watt-meyer2023} report performance variations of up to a factor of four with different training seeds for SFNO, but we observe that rollouts with all tested architectures completely degenerate for some random seeds. We point out that our results were obtained on a relatively coarse resolution of \SI{5.625}{\degree} and it is unclear if they generalize to higher spatial resolutions.
While we focus on stability as a necessary criterion for autoregressive DL climate models, we acknowledge that evaluating the overall quality of a climate model is much more complex.

\section{Conclusion \& Future Work}

With this study, we provide insights into basic hyperparameter choices for training DL models on reanalysis data which help to build autoregressive atmospheric models suitable for climate timescales. While models based on the geometry-aware SFNO architecture yield stable rollouts across many tested hyperparameter configurations, we find that non geometry-aware architectures can yield rollouts of equal stability with the right hyperparameter settings. Furthermore, we show that the multi-step objective helps to stabilize long rollouts across all architectures. The sensitivity to the training random seed and the used set of prognostic variables illustrates that further empirical and theoretical analysis is needed to create a more profound understanding of criteria that influence stability of long rollouts. Future work should continue to study the design choices which might impact the behavior of long autoregressive rollouts, e.g., optimization, regularization techniques and the number of input/output timesteps. While long-term stability of autoregressive atmospheric DL models is a prerequisite for achieving fully data-driven climate projections, integrating the external forcings that lead to changing climate and the coupling to other earth system components are necessary steps towards this goal.

\section*{Acknowledgement}
This research was conducted in the BigData@Geo 2.0 project which is co-financed by the European Regional Development Fund (ERDF).
The authors gratefully acknowledge the scientific support and HPC resources provided by the Erlangen National High Performance Computing Center (NHR@FAU) of the Friedrich-Alexander-Universität Erlangen-Nürnberg (FAU) under the NHR project ID b214cb. 
NHR funding is provided by federal and Bavarian state authorities.
NHR@FAU hardware is partially funded by the German Research Foundation (DFG) – 440719683.

\bibliography{iclr2025_conference.bib}
\bibliographystyle{iclr2025_conference.bst}
\newpage
\appendix
\section{Parameter Settings}
 We train each model configuration for 20 epochs at float32 precision and employ early stopping when the validation loss has not improved during the last 5 epochs. We use the Adam optimizer with an initial learning rate of \SI{4e-3} for FCN and ClimaX, and a learning rate of \SI{1e-3} for SFNO models. A smaller LR for SFNO was necessary as the multi-step training loss exploded at the start of training with the LR of \SI{4e-3}. The LR rate was chosen based on the training protocol of \citet{karlbauer2024a} accounting for training with the 4 times larger effective batch size of 64. The learning rate decays every epoch with a cosine annealing schedule towards 0. We clip the norm of gradients to a value of \SI{0.001} similar to \citet{karlbauer2024a}. The following random seeds were used for training: 597, 1152, 1826, 3909, 6153, 5513, 5707, 9813, 9941, 9982. Model specific hyperparameters are listed in \cref{tab:hyperparameters}.

\begin{table}[h]
    \renewcommand{\arraystretch}{1.2}
    \centering
    \caption{Model Specific Hyperparameters}
    \begin{tabular}{|l|l|l|}
    \toprule
    \multicolumn{1}{|c|}{ClimaX} & \multicolumn{1}{|c|}{FourCastNet} & \multicolumn{1}{|c|}{SFNO}\\
    \midrule
    Patch size: (2,2) & Patch size: (1,1) & Scale Factor: 1\\
    \# Heads: 8& \# Blocks: 4 &  Big Skip: False\\
    Decoder Depth: 2 & Use Pos. Embed.: False & Use Pos. Embed.: False \\
    MLP Ratio: 4 & MLP Ratio: 4 & Use MLP: True \\
    Drop Rate: 0.0 & Drop Rate: 0.0 & Grid: "Equiangular" \\
    Drop Path: 0.0 & Drop Path: 0.0 & factorization: Null \\
    & Hard Threshold Fraction: 1.0 &  Hard Threshold Fraction: 1.0\\
    & Sparsity Threshold: 0.01 & \\
    \bottomrule
    \end{tabular}
    \label{tab:hyperparameters}
\end{table}

\begin{table}[h]
    \centering
    \setlength{\tabcolsep}{5pt}
    \caption{List of prognostic, constant, and forcing variables used as input variables.}
    \label{tab:variables}
    \begin{tabular}{llll}
        \toprule
        \textbf{Category} & \textbf{Vars} & \textbf{Level} & \textbf{Description} \\
        \midrule
        \multirow{8}{*}{\parbox{2cm}{Prognostic (33 Variables)}} 
        & tas & 2 m & Near-surface temperature\\
        & uas & 10 m & Near-surface eastward wind comp.\\
        & vas & 10 m & Near-surface northward wind comp.\\
        & ta & 925, 850, 700, 600, 500, 250 hPa & Temperature\\
        & zg & 925, 850, 700, 600, 500, 250 hPa & Geopotential height\\
        & hus & 925, 850, 700, 600, 500, 250 hPa & Specific humidity\\
        & ua & 925, 850, 700, 600, 500, 250 hPa & Eastward wind comp.\\
        & va & 925, 850, 700, 600, 500, 250 hPa & Northward wind comp.\\
        \midrule
        \multirow{5}{*}{\parbox{1.8cm}{Prognostic (8 Variables)}} & tas & 2 m & Near-surface temperature \\
        & uas & 10 m & Near-surface eastward wind comp. \\
        & vas & 10 m & Near-surface northward wind comp. \\
        & ta & 850 hPa & Temperature \\
        & zg & 1000, 700, 500, 300 hPa & Geopotential height \\
        \midrule
        \multirow{4}{*}{Constant} & lsm & - & Binary land-sea representation \\
        & orog & - & Surface elevation data \\
        & lat & - & Latitude information as 2D grid\\
         & lon & - & Longitude information as 2D grid \\
        \midrule
        \multirow{1}{*}{Forcing} & tisr & - & \parbox{4cm}{Top-of-the-atmosphere incident solar radiation} \\
        \bottomrule
    \end{tabular}
\end{table}

\section{Additional Figures}

\begin{figure}[h]

\includegraphics[width=\textwidth]{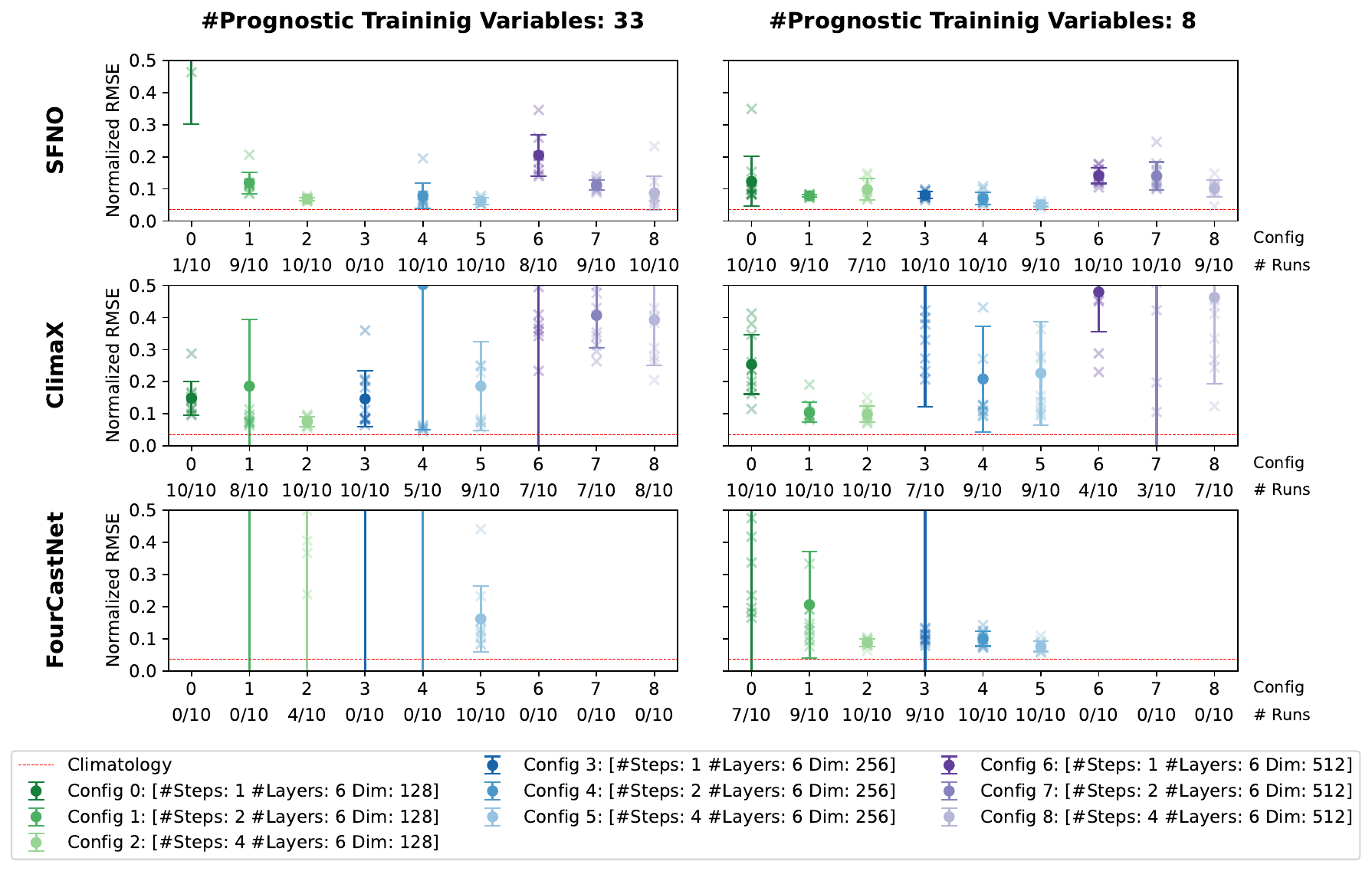}
\caption{As \cref{fig:overview-layers4} with $L=6$: Area-weighted normalized RMSE scores averaged over the variables \texttt{tas}, \texttt{uas}, \texttt{vas}, \texttt{ta850} and \texttt{zg500} for model configurations $0-8$ with layer size $L=6$. The mean (dots) and standard deviation (error bars) were computed across 10 seeds (crosses) with finite RMSE. For better readability, the y-axis is cut-off at 0.5 and the number of displayed runs out of 10 is shown on the x-axis.}
\label{fig:overview-layers6}
\end{figure}

\begin{figure}[h]
\includegraphics[width=\textwidth]{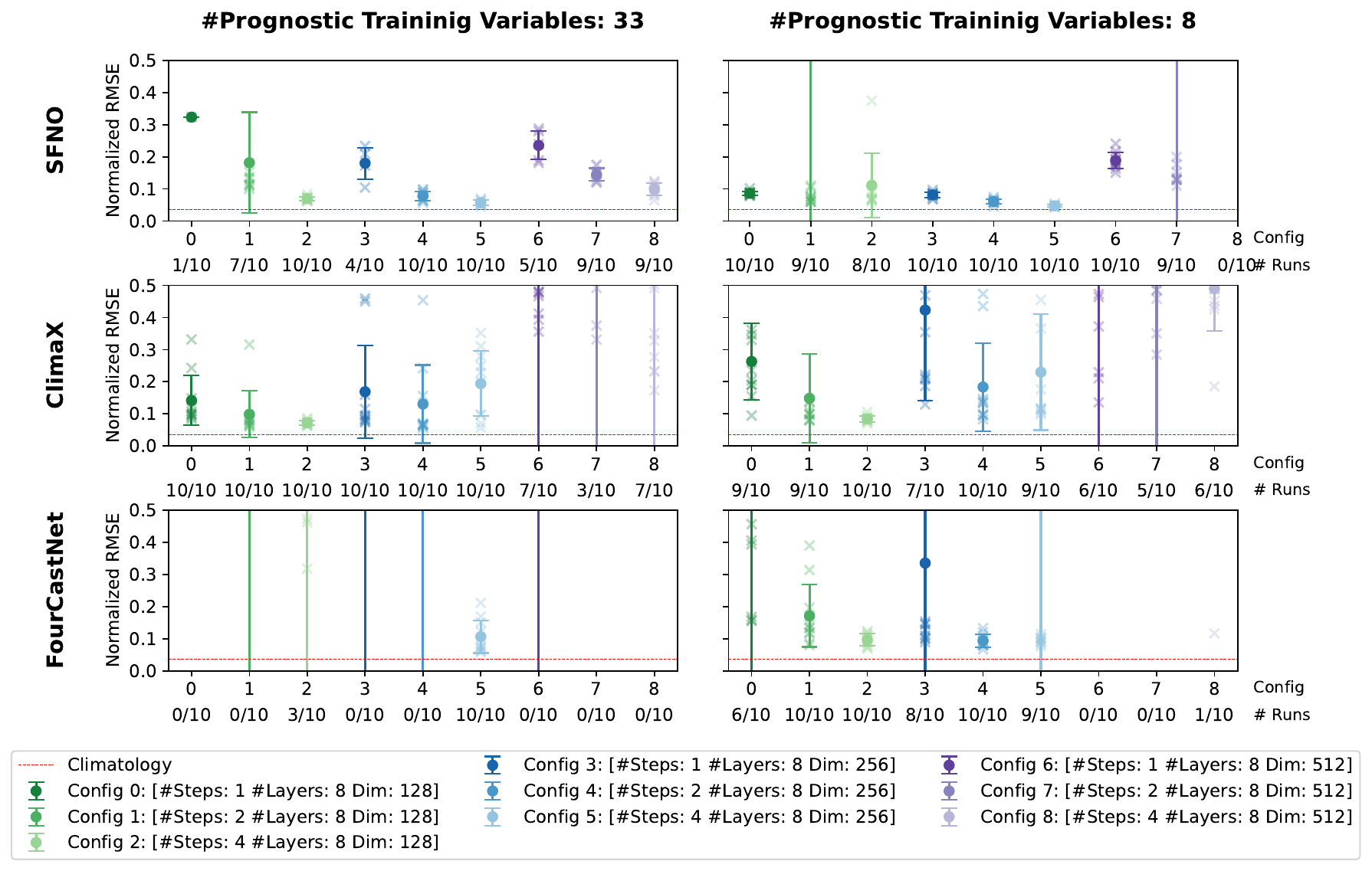}
\caption{As \cref{fig:overview-layers4} with $L=8$: Area-weighted normalized RMSE scores averaged over the variables \texttt{tas}, \texttt{uas}, \texttt{vas}, \texttt{ta850} and \texttt{zg500} for model configurations $0-8$ with layer size $L=8$. The mean (dots) and standard deviation (error bars) were computed across 10 seeds (crosses) with finite RMSE. For better readability, the y-axis is cut-off at 0.5 and the number of displayed runs out of 10 is shown on the x-axis.}
\label{fig:overview-layers8}
\end{figure}

\begin{figure}[h]
\includegraphics[width=\textwidth]{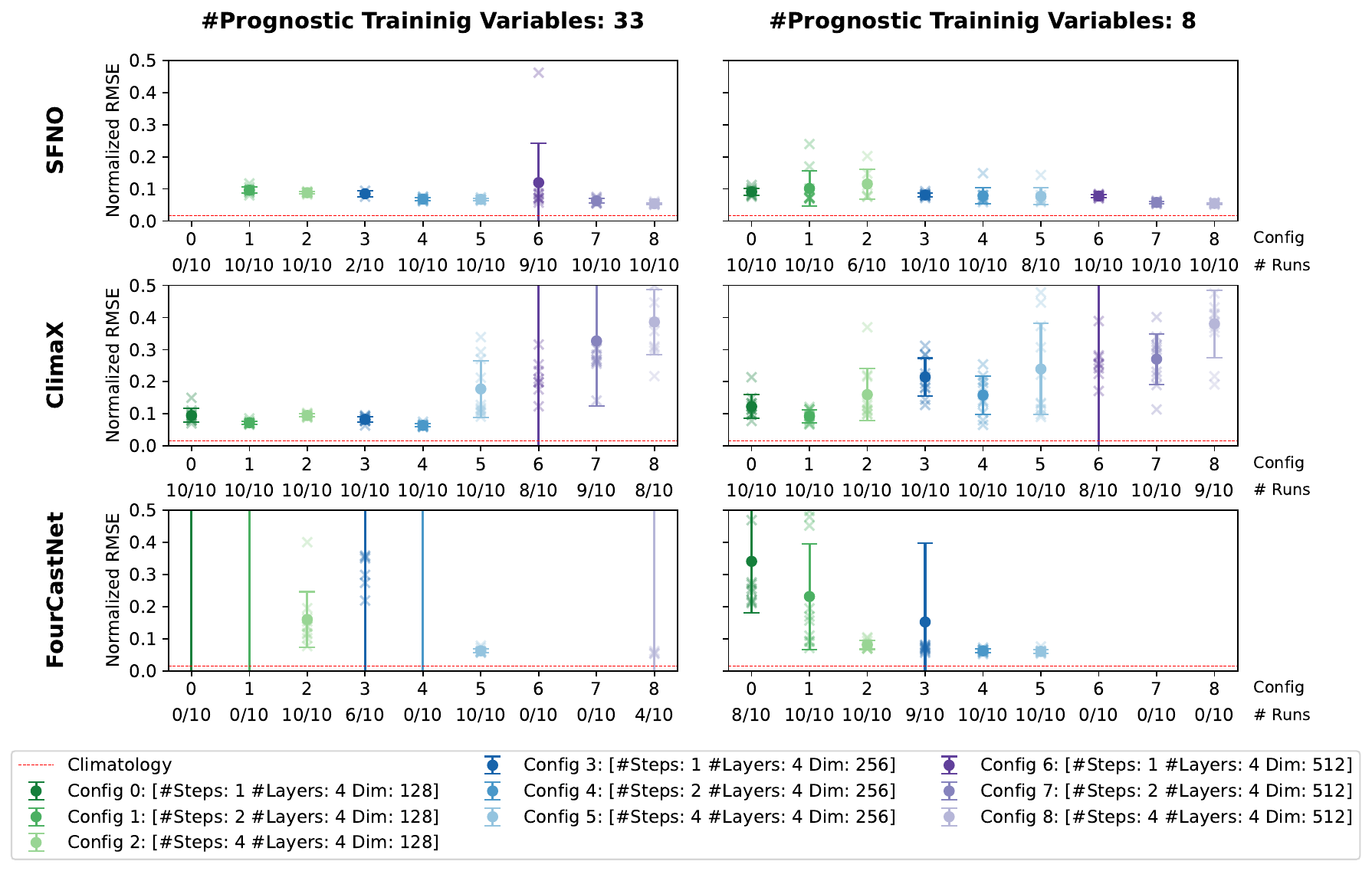}
\caption{As \cref{fig:overview-layers4}, but temporal standard deviations are compared instead of temporal means: Area-weighted normalized RMSE scores for the temporal standard deviation averaged over the variables \texttt{tas}, \texttt{uas}, \texttt{vas}, \texttt{ta850} and \texttt{zg500} for model configurations $0-8$ with layer size $L=4$. The mean (dots) and standard deviation (error bars) were computed across 10 seeds (crosses) with finite RMSE. For better readability, the y-axis is cut-off at 0.5 and the number of displayed runs out of 10 is shown on the x-axis.}
\label{fig:std-layers6}
\end{figure}

\begin{figure}[h]
\includegraphics[width=\textwidth]{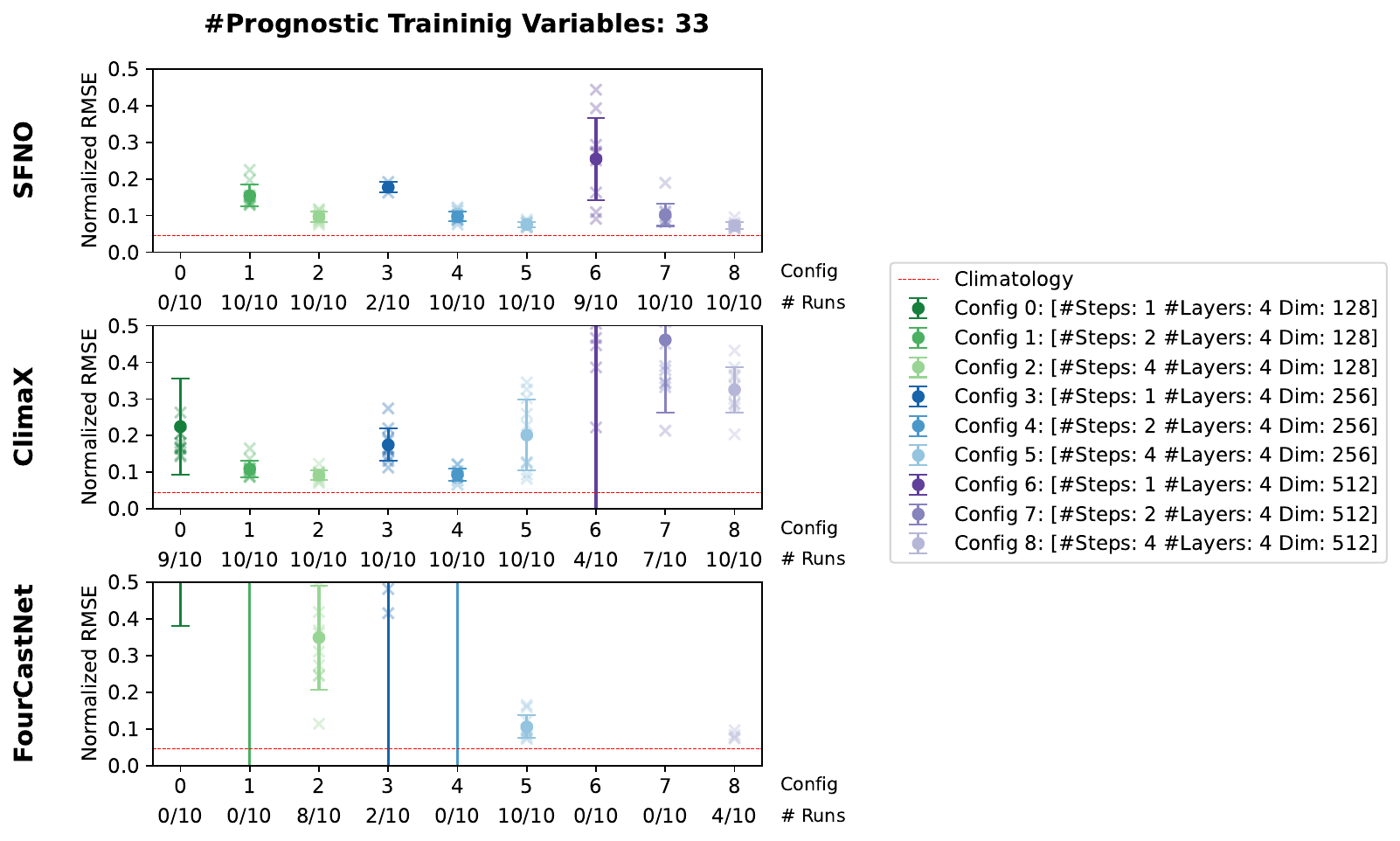}
\caption{As \cref{fig:overview-layers4} averaged over all 33 variables. Area-weighted normalized RMSE scores averaged over all 33 prognostic variables for model configurations $0-8$ with layer size $L=4$. The mean (dots) and standard deviation (error bars) were computed across 10 seeds (crosses) with finite RMSE. For better readability, the y-axis is cut-off at 0.5 and the number of displayed runs out of 10 is shown on the x-axis.}
\label{fig:33vars-layers6}
\end{figure}


\begin{figure}[h]
\centering
\includegraphics[width=\textwidth]{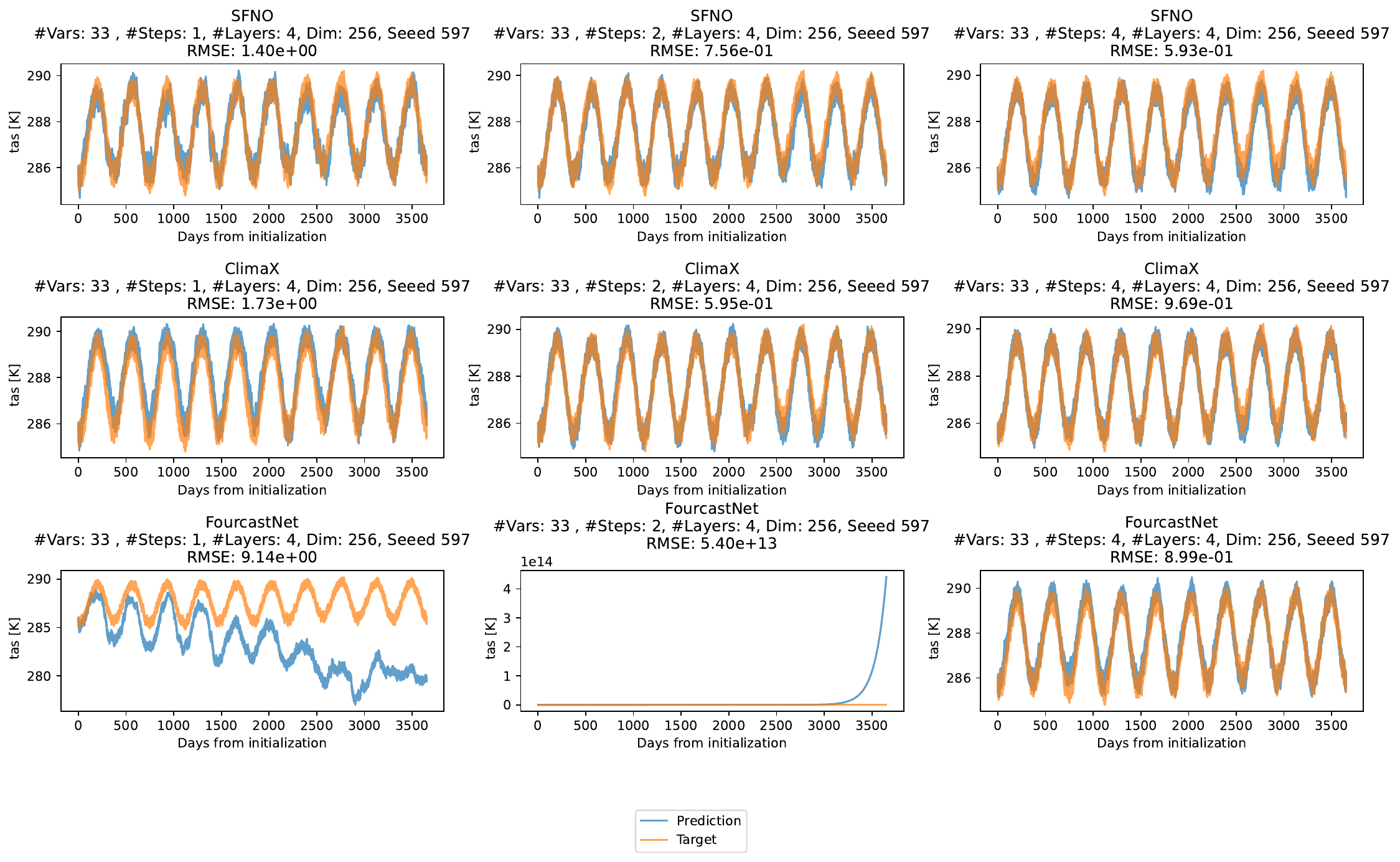}
\caption{Near-surface temperature timeseries for selected models and ERA5 data that show the area-weighted global means for the 10-year rollout period. The RMSE score for the shown variable (not normalized) is indicated above each plot. }
\label{fig:model-timeseries}
\end{figure}

\begin{figure}[h]
\centering
\includegraphics[width=\textwidth]{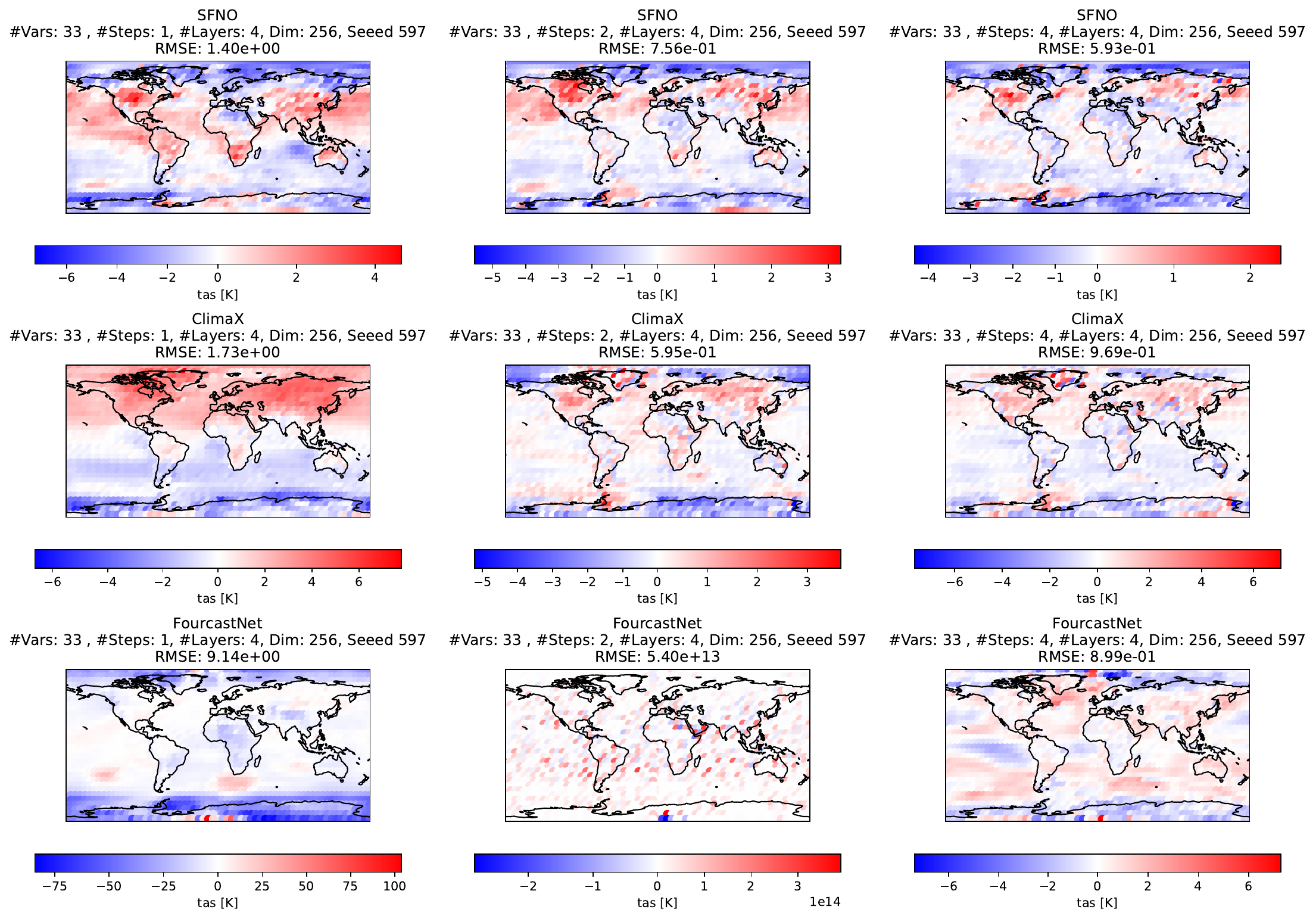}
\caption{Near-surface temperature maps for selected models that show the difference of the temporal means of the 10-year rollouts minus the the temporal mean of ERA5 data. The RMSE score for the shown variable (not normalized) is indicated above each plot.}
\label{fig:model-maps}
\end{figure}

\begin{figure}[h]
\centering
\includegraphics[width=\textwidth]{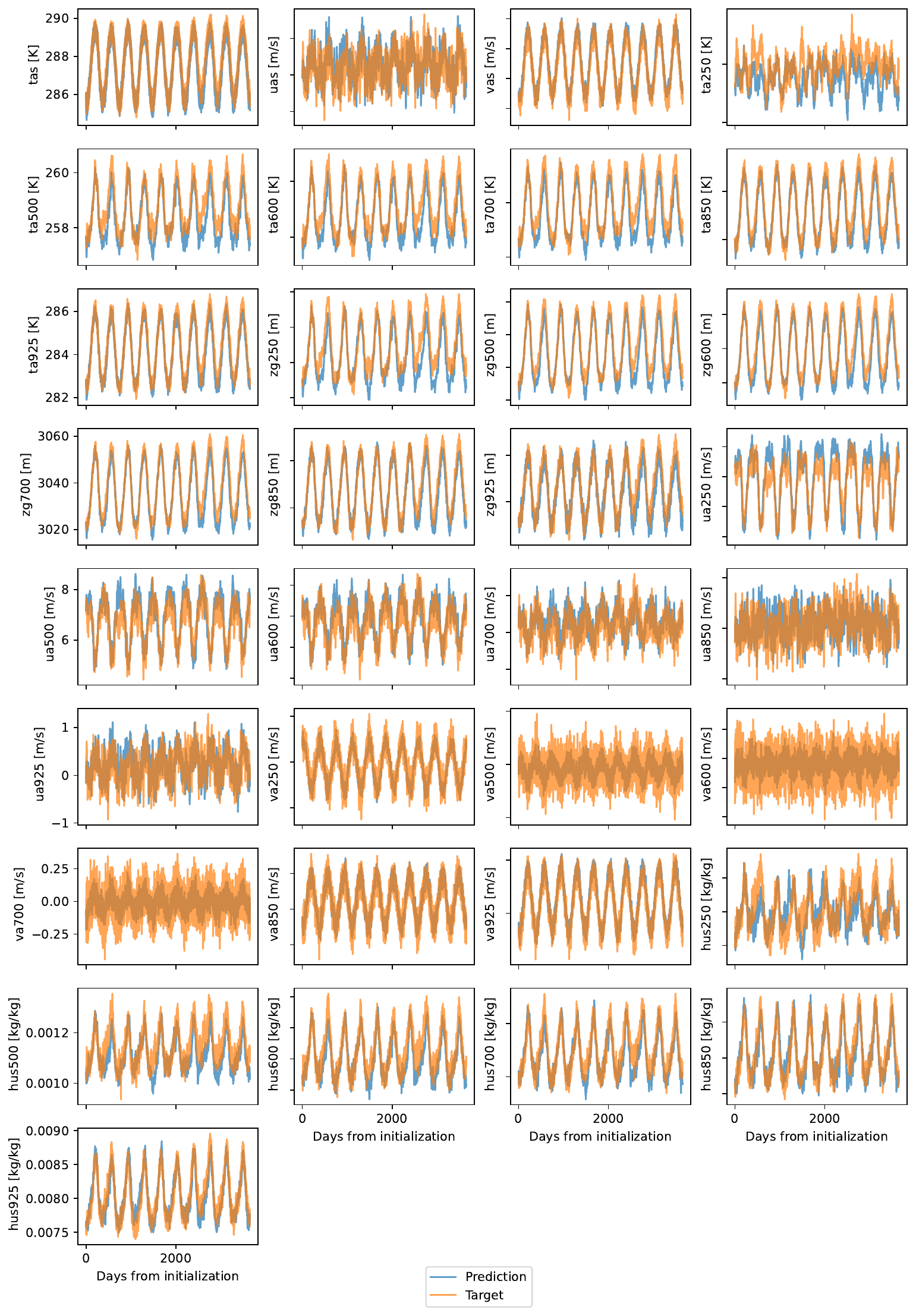}
\caption{Timeseries of all variables for the SFNO model with 4-steps autoregressive training, 4 layers, hidden dimension of 512 and random seed 597 that show the area-weighted global means for the 10-year rollout.}
\label{fig:sfno-rmse-per-var}
\end{figure}

\begin{figure}[h]
\centering
\includegraphics[height=0.9\textheight]{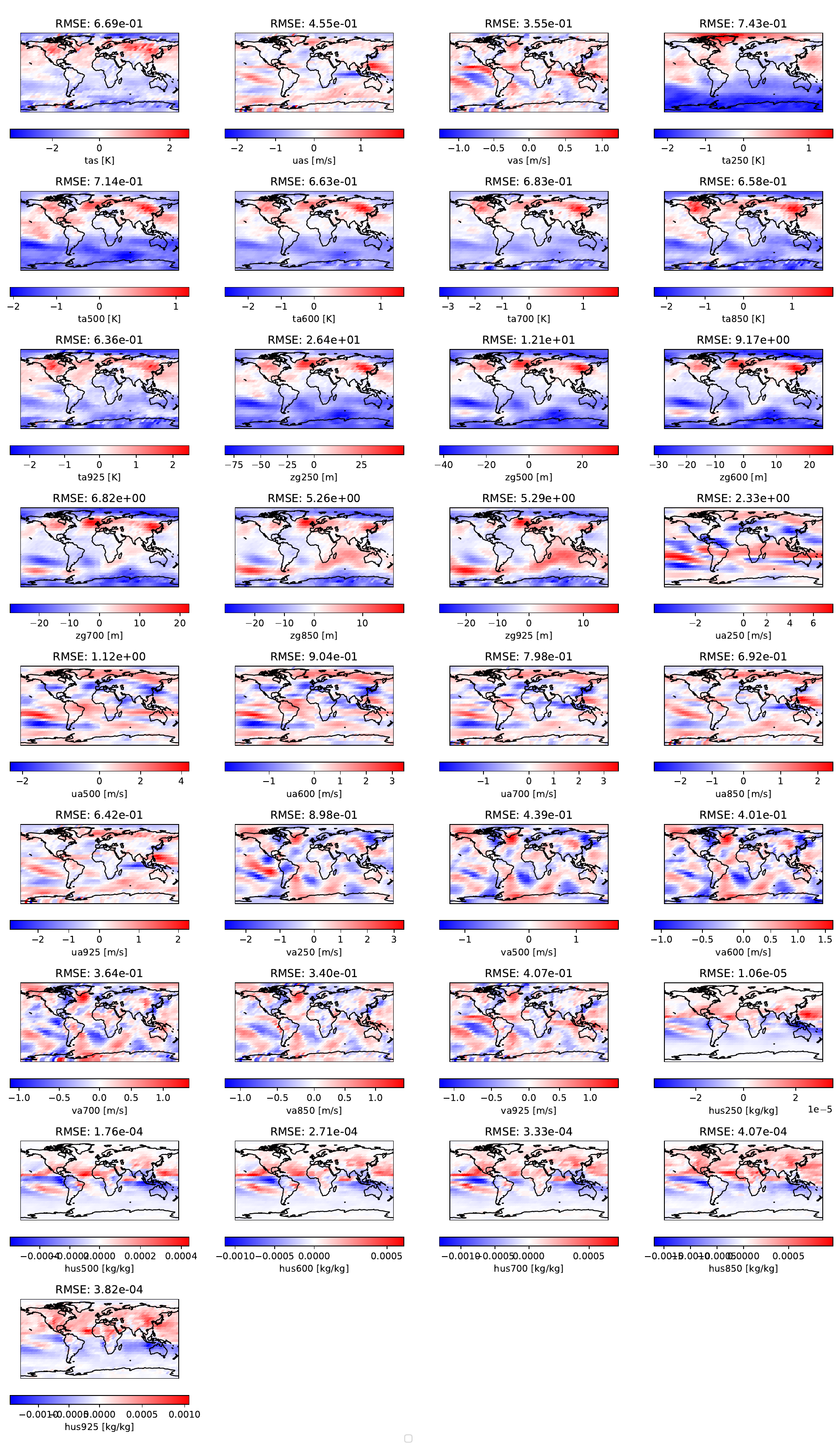}
\caption{
Maps of all variables for the SFNO model with 4-step autoregressive training, 4 layers, hidden dimension of 512 and random seed 597 that show the difference of the temporal means of the 10-year rollout minus the temporal mean of ERA5 data. The RMSE score for the shown variable (not normalized) is indicated above each plot.}
\label{fig:sfno-maps-per-var}
\end{figure}

\end{document}

%% file: math_commands.tex
\usepackage{amsmath,amsfonts,bm}

\def\eqref#1{equation~\ref{#1}}

\def\1{\bm{1}}

\DeclareMathAlphabet{\mathsfit}{\encodingdefault}{\sfdefault}{m}{sl}
\SetMathAlphabet{\mathsfit}{bold}{\encodingdefault}{\sfdefault}{bx}{n}

